\def\year{2019}\relax
\documentclass[letterpaper]{article} 
\usepackage{aaai19}  
\usepackage{times}  
\usepackage{helvet}  
\usepackage{courier}  
\usepackage{url}  
\usepackage{graphicx}  
\usepackage{amsmath, amssymb, amsthm, amsfonts}
\usepackage{booktabs}
\usepackage{multirow}
\usepackage{subcaption}

\newtheorem{definition}{Definition}

\nocopyright

\begin{document}
%
\title{Generating Artificial Data for Private Deep Learning}
\author{Aleksei Triastcyn \and Boi Faltings\\
	Artificial Intelligence Laboratory \\
	Ecole Polytechnique F\'ed\'erale de Lausanne \\
	Lausanne, Switzerland \\
	\texttt{\{aleksei.triastcyn, boi.faltings\}@epfl.ch} \\
}

\maketitle

\begin{abstract}
In this paper, we propose generating artificial data that retain statistical properties of real data as the means of providing privacy for the original dataset. We use generative adversarial networks to draw privacy-preserving artificial data samples and derive an empirical method to assess the risk of information disclosure in a differential-privacy-like way. Our experiments show that we are able to generate labelled data of high quality and use it to successfully train and validate supervised models. Finally, we demonstrate that our approach significantly reduces vulnerability of such models to model inversion attacks.
\end{abstract}

\section{Introduction}
\label{sec:introduction}

Following recent advancements in deep learning, more and more people and companies get interested in putting their data in use and employ machine learning (ML) to generate a wide range of benefits that span financial, social, medical, security, and other aspects. At the same time, however, such models are able to capture a fine level of detail in training data, potentially compromising privacy of individuals whose features sharply differ from others. 
Recent research~\cite{fredrikson2015model} suggests that even without access to internal model parameters it is possible to recover (up to a certain degree) individual examples, e.g. faces, from the training set.

The latter result is especially disturbing knowing that deep learning models are becoming an integral part of our lives, making its way to phones, smart watches, cars, and appliances. And since these models are often trained on customers' data, such training set recovery techniques endanger privacy even without access to the manufacturer's servers where these models are being trained.

One direction to tackle this problem is enforcing privacy during training~\cite{abadi2016deep,papernot2016semi,papernot2018scalable}. We will refer to these techniques as \emph{model release} methods. While these approaches perform well in ML tasks and provide strong privacy guarantees, they are often restrictive. First and foremost, releasing a single trained model does not provide much flexibility in the future. For instance, it would significantly reduce possibilities for combining models trained on data from different sources. Evaluating a variety of such models and picking the best one is also complicated by the need of adjusting private training for each of them.
Moreover, most of these methods assume (implicitly or explicitly) access to public data of similar nature, which may not be possible in areas like medicine.

\begin{figure}
	\centering
	\includegraphics[width=\linewidth]{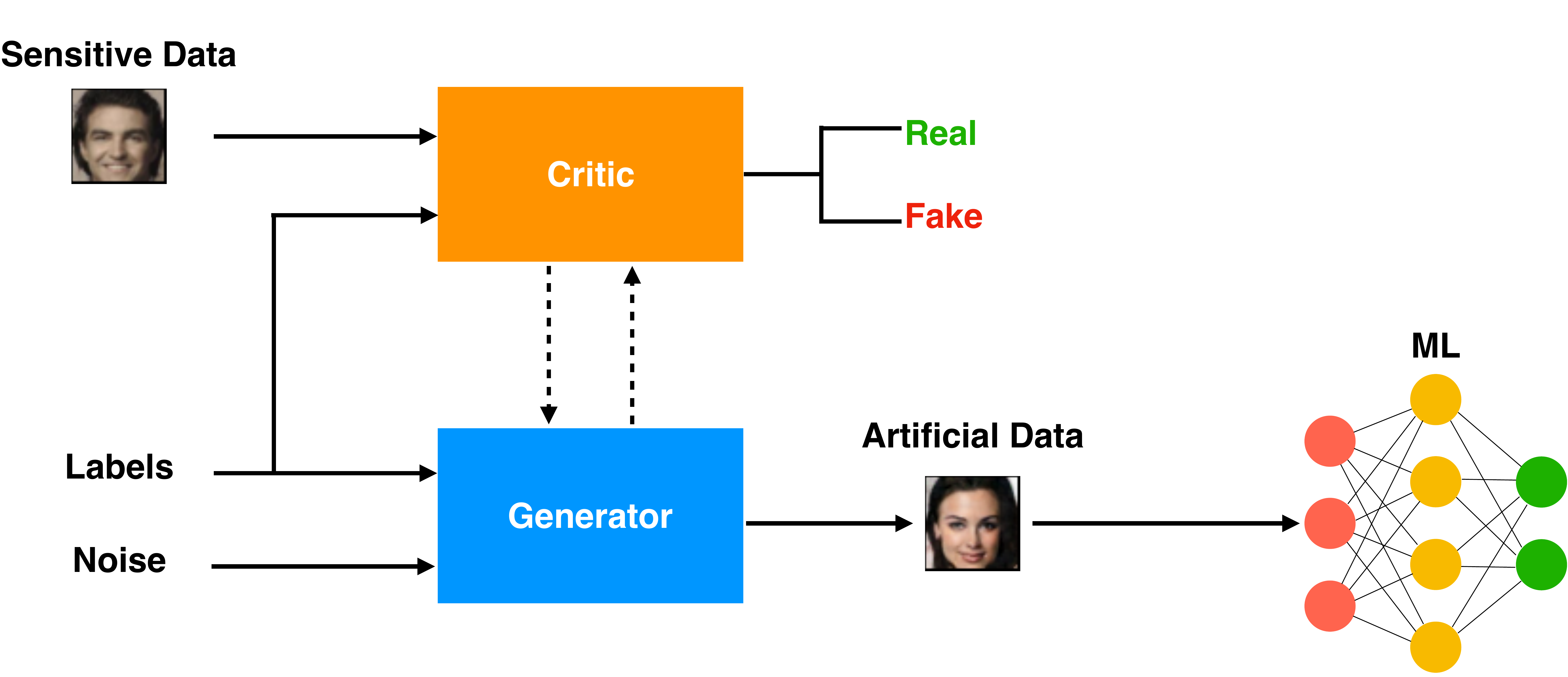}
    	\caption{Architecture of our solution. Sensitive data is used to train a GAN to produce a private artificial dataset, which then can be used by any ML model.}
	\label{fig:architecture}
\end{figure}

In contrast, we study the task of privacy-preserving \emph{data release}, which has many immediate advantages. First, any ML model could be trained on released data without additional assumptions. Second, data from different sources could be easily pooled to build stronger models. Third, released data could be traded on data markets\footnote{\url{https://www.datamakespossible.com/value-of-data-2018/dawn-of-data-marketplace}}, where anonymisation and protection of sensitive information is one of the biggest obstacles. Finally, data publishing would facilitate transparency and reproducibility of research studies.

In particular, we are interested in solving two problems. First, how to preserve high utility of data for ML algorithms while protecting sensitive information in the dataset. Second, how to quantify the risk of recovering private information from the published dataset, and thus, the trained model. 

The main idea of our approach is to use generative adversarial networks (GANs)~\cite{goodfellow2014generative} to create artificial datasets to be used in place of real data for training. This method has a number of advantages over the earlier work \cite{abadi2016deep,papernot2016semi,papernot2018scalable,bindschaedler2017plausible}. First of all, our solution allows releasing entire datasets, thereby possessing all the benefits of private \emph{data release} as opposed to \emph{model release}. Second, it achieves high accuracy without pre-training on similar public data. Third, it is more intuitive and flexible, e.g. it does not require a complex distributed architecture.

To estimate potential privacy risks, we design an \emph{ex post} analysis framework for generated data. We use KL divergence estimation and Chebyshev's inequality to find statistical bounds on expected privacy loss for a dataset in question.

Our contributions in this paper are the following:
\begin{itemize}
\item we propose a novel, yet simple, approach for private data release, and to the best of our knowledge, this is the first practical solution for complex real-world data;
\item we introduce a new framework for statistical estimation of potential privacy loss of the released data;
\item we show that our method achieves learning performance of model release methods and is resilient to model inversion attacks.
\end{itemize}

The rest of the paper is structured as follows. In Section~\ref{sec:related_work}, we give an overview of related work. Section~\ref{sec:preliminaries} contains some preliminaries. In Section~\ref{sec:approach}, we describe our approach and privacy estimation framework, and discuss its limitations. Experimental results and implementation details are presented in Section~\ref{sec:evaluation}, and Section~\ref{sec:conclusion} concludes the paper.

\section{Related Work}
\label{sec:related_work}

In recent years, as machine learning applications become a commonplace, a body of work on security of these methods grows at a rapid pace. Several important vulnerabilities and corresponding attacks on ML models have been discovered, raising the need of devising suitable defences. Among the attacks that compromise privacy of training data, model inversion~\cite{fredrikson2015model} and membership inference~\cite{shokri2017membership} received high attention.

Model inversion~\cite{fredrikson2015model} is based on observing the output probabilities of the target model for a given class and performing gradient descent on an input reconstruction. Membership inference~\cite{shokri2017membership} assumes an attacker with access to similar data, which is used to train a "shadow" model, mimicking the target, and an attack model. The latter predicts if a certain example has already been seen during training based on its output probabilities. Note that both attacks can be performed in a black-box setting, without access to the model internal parameters.

To protect privacy while still benefiting from the use of statistics and ML, many techniques have been developed over the years, including $k$-anonymity~\cite{sweeney2002}, $l$-diversity~\cite{machanavajjhala2007}, $t$-closeness~\cite{li2007t}, and differential privacy (DP)~\cite{dwork2006}. The latter has been recognised as a rigorous standard and is widely accepted by the research community. Its generic formulation, however, makes it hard to achieve and to quantify potential privacy loss of the already trained model. To overcome this, we build upon notions of empirical DP~\cite{abowd2013differential} and on-average KL privacy~\cite{wang2016average}.

Most of the ML-specific literature in the area concentrates on the task of privacy-preserving model release. One take on the problem is to distribute training and use disjoint datasets. For example,  \citeauthor{shokri2015privacy}~\shortcite{shokri2015privacy} propose to train a model in a distributed manner by communicating sanitised updates from participants to a central authority. Such a method, however, yields high privacy losses~\cite{abadi2016deep,papernot2016semi}. An alternative technique suggested by \citeauthor{papernot2016semi}~\shortcite{papernot2016semi}, also uses disjoint training sets and builds an ensemble of independently trained teacher models to transfer knowledge to a student model by labelling public data. This result has been extended in \cite{papernot2018scalable} to achieve state-of-the-art image classification results in a private setting (with single-digit DP bounds).
A different approach is taken by \citeauthor{abadi2016deep}~\shortcite{abadi2016deep}. They suggest using differentially private stochastic gradient descent (DP-SGD) to train deep learning models in a private manner. This approach achieves high accuracy while maintaining low DP bounds, but may also require pre-training on public data.

A more recent line of research focuses on private data release and providing privacy via generating synthetic data~\cite{bindschaedler2017plausible,huang2017context,beaulieu2017privacy}. In this scenario, DP is hard to guarantee, and thus, such models either relax the DP requirements or remain limited to simple data. In \cite{bindschaedler2017plausible}, authors use a graphical probabilistic model to learn an underlying data distribution and transform real data points (seeds) into synthetic data points, which are then filtered by a privacy test based on a \emph{plausible deniability} criterion. This procedure would be rather expensive for complex data, such as images. \citeauthor{huang2017context}~\shortcite{huang2017context} introduce the notion of \emph{generative adversarial privacy} and use GANs to obfuscate real data points w.r.t. pre-defined private attributes, enabling privacy for more realistic datasets. Finally, a natural approach to try is training GANs using DP-SGD~\cite{beaulieu2017privacy,xie2018differentially,zhang2018differentially}. However, it proved extremely difficult to stabilise training with the necessary amount of noise, which scales as $\sqrt{m}$ w.r.t. the number of model parameters $m$. It makes these methods inapplicable to more complex datasets without resorting to unrealistic (at least for some areas) assumptions, like access to public data from the same distribution.

Similarly, our approach uses GANs, but data is generated without real seeds or applying noise to gradients. Instead, we verify experimentally that out-of-the-box GAN samples can be sufficiently different from real data, and expected privacy loss is empirically bounded by single-digit numbers.

\section{Preliminaries}
\label{sec:preliminaries}

This section provides necessary definitions and background. Let us commence with approximate differential privacy.

\begin{definition}
A randomised function (mechanism) $\mathcal{M}: \mathcal{D} \rightarrow \mathcal{R}$ with domain $\mathcal{D}$ and range $\mathcal{R}$ satisfies $(\varepsilon, \delta)$-differential privacy if for any two adjacent inputs $d, d' \in \mathcal{D}$ and for any outcome $o \in \mathcal{R}$ the following holds:
\begin{align}
	\Pr\left[\mathcal{M}(d)=o\right] \leq e^\varepsilon \Pr\left[\mathcal{M}(d') = o\right] + \delta.
\end{align}
\end{definition}

\begin{definition}
Privacy loss of a randomised mechanism $\mathcal{M}: \mathcal{D} \rightarrow \mathcal{R}$ for inputs $d, d' \in \mathcal{D}$ and outcome $o \in \mathcal{R}$ takes the following form:
\begin{align}
	L_{(\mathcal{M}(d) \| \mathcal{M}(d'))} = \log\frac{\Pr\left[\mathcal{M}(d) = o \right]}{\Pr\left[\mathcal{M}(d') = o \right]}.
\end{align}
\end{definition}

\begin{definition}
The Gaussian noise mechanism achieving $(\varepsilon, \delta)$-DP, for a function $f: \mathcal{D} \rightarrow \mathbb{R}^m$, is defined as
\begin{align}
	\mathcal{M}(d) = f(d) + \mathcal{N}(0, \sigma^2),
\end{align}
where $\sigma > C \sqrt{2\log\frac{1.25}{\delta}} / \varepsilon$ and $C$ is the L2-sensitivity of $f$.
\end{definition}

For more details on differential privacy and the Gaussian mechanism, we refer the reader to~\cite{dwork2014algorithmic}.

In our privacy estimation framework, we also use some classical notions from probability and information theory.

\begin{definition}
The Kullback–Leibler (KL) divergence between two continuous probability distributions $P$ and $Q$ with corresponding densities $p$, $q$ is given by:
\begin{align}
	D_{KL}(P \| Q) = \int_{-\infty}^{+\infty} p(x) \log\frac{p(x)}{q(x)} dx.
\end{align}
\end{definition}

Note that KL divergence between the distributions of $\mathcal{M}(d)$ and $\mathcal{M}(d')$ is nothing but the expectation of the privacy loss random variable $\mathbb{E}[L_{(\mathcal{M}(d) \| \mathcal{M}(d'))}]$.

Finally, Chebyshev's inequality is used to obtain tail bounds. In particular, as we expect the distribution to be asymmetric, we use the version with semi-variances~\cite{berck1982using} to get a sharper bound:
\begin{align}
\label{eq:chebyshev}
	\Pr(x \geq \mathbb{E}[x] + k\sigma) \leq \frac{1}{k^2} \frac{\sigma_{+}^2}{\sigma^2},
\end{align}
where $\sigma_{+}^2 = \int_{\mathbb{E}[x]}^{+\infty} p(x) (x - \mathbb{E}[x])^2 dx$ is the upper semi-variance.

\section{Our Approach}
\label{sec:approach}

In this section, we describe our solution, its further improvements, and provide details of the privacy estimation framework. We then discuss limitations of the method. More background on privacy can be found in~\cite{dwork2014algorithmic}.

The main idea of our approach is to use artificial data for learning and publishing instead of real (see Figure~\ref{fig:architecture} for a general workflow). The intuition behind it is the following. Since it is possible to recover training examples from ML models~\cite{fredrikson2015model}, we need to limit the exposure of real data during training. While this can be achieved by DP training (e.g. DP-SGD), it would have the limitations mentioned earlier. Moreover, certain attacks can still be successful if DP bounds are loose~\cite{hitaj2017deep}. Removing real data from the training process altogether would add another layer of protection and limit the information leakage to artificial samples. What remains to show is that artificial data is sufficiently different from real.

\subsection{Differentially Private Critic}

Despite the fact that the generator does not have access to real data in the training process, one cannot guarantee that generated samples will not repeat the input. To alleviate this problem, we propose to enforce differential privacy on the output of the discriminator (\emph{critic}). This is done by employing the Gaussian noise mechanism~\cite{dwork2014algorithmic} at the second-to-last layer: clipping the $L2$ norm of the input and adding Gaussian noise. To be more specific, activations $a(x)$ of the second-to-last layer become $\tilde{a}(x) = a(x) / \max(\|a(x)\|_2, 1) + \mathcal{N}(0; \sigma^2)$. We refer to this version of the critic as \emph{DP critic}.

Note that if the chosen GAN loss function was directly differentiable w.r.t. generator output, i.e. if critic could be treated as a black box, this modification would enforce the same DP guarantees on generator parameters, and consequently, all generated samples. Unfortunately, this is not the case for practically all existing versions of GANs, including WGAN-GP~\cite{gulrajani2017improved} used in our experiments.

As our evaluation shows, this modification has a number of advantages. First, it improves diversity of samples and decreases similarity with real data. Second, it allows to prolong training, and hence, obtain higher quality samples. Finally, in our experiments, it significantly improves the ability of GANs to generate samples conditionally.

\subsection{Privacy Estimation Framework}

Our framework builds upon ideas of \emph{empirical DP} (EDP)~\cite{abowd2013differential,schneider2015new} and \emph{on-average KL privacy}~\cite{wang2016average}. The first can be viewed as a measure of sensitivity on posterior distributions of outcomes~\cite{charest2017meaning} (in our case, generated data distributions), while the second relaxes DP notion to the case of an average user.

As we don't have access to exact posterior distributions, a straightforward EDP procedure in our scenario would be the following:
\emph{(1)} train GAN on the original dataset $D$; 
\emph{(2)} remove a random sample from $D$; 
\emph{(3)} re-train GAN on the updated set; 
\emph{(4)} estimate probabilities of all outcomes and the maximum privacy loss value; 
\emph{(5)} repeat \emph{(1)--(4)} sufficiently many times to approximate $\varepsilon$, $\delta$.

If the generative model is simple, this procedure can be used without modification. Otherwise, for models like GANs, it becomes prohibitively expensive due to repetitive re-training (steps \emph{(1)--(3)}). Another obstacle is estimating the maximum privacy loss value (step \emph{(4)}). To overcome these two issues, we propose the following.

First, to avoid re-training, we imitate the removal of examples directly on the generated set $\widetilde{D}$. We define a similarity metric $sim(x, y)$ between two data points $x$ and $y$ that reflects important characteristics of data (see Section~\ref{sec:evaluation} for details). For every randomly selected real example $i$, we remove $k$ nearest artificial neighbours to simulate absence of this example in the training set and obtain $\widetilde{D}^{-i}$. Our intuition behind this operation is the following. Removing a real example would result in a lower probability density in the corresponding region of space. If this change is picked up by a GAN, which we assume is properly trained (e.g. there is no mode collapse), the density of this region in the generated examples space should also decrease. The number of neighbours $k$ is a hyper-parameter. In our experiments, it is chosen heuristically by computing KL divergence between the real and artificial data distributions and assuming that all the difference comes from one point.

Second, we propose to relax the worst-case privacy loss bound in step \emph{(4)} by the expected-case bound, in the same manner as on-average KL privacy. This relaxation allows us to use a high-dimensional KL divergence estimator~\cite{perez2008kullback} to obtain the expected privacy loss for every pair of adjacent datasets ($\widetilde{D}$ and $\widetilde{D}^{-i}$). There are two major advantages of this estimator: it converges almost surely to the true value of KL divergence; and it does not require intermediate density estimates to converge to the true probability measures. Also since this estimator uses nearest neighbours to approximate KL divergence, our heuristic described above is naturally linked to the estimation method.

Finally, after obtaining sufficiently many samples of different pairs $(\widetilde{D}, \widetilde{D}^{-i})$, we use Chebyshev's inequality to bound the probability $\gamma = \Pr(\mathbb{E}[L_{(\mathcal{M}(D) \| \mathcal{M}(D'))}] \geq \mu)$ of the expected privacy loss~\cite{dwork2016concentrated} exceeding a predefined threshold $\mu$. To deal with the problem of insufficiently many samples, one could use a sample version of inequality~\cite{saw1984chebyshev} at the cost of looser bounds.

\subsection{Limitations}

Our empirical privacy estimator could be improved in a number of ways. For instance, providing worst-case privacy loss bounds would be largely beneficial. Furthermore, simulating the removal of training examples currently depends on heuristics and the chosen similarity metric, which may not lead to representative samples and therefore, poor guarantees.

We provide bounds on expected privacy loss based on \emph{ex post} analysis of the artificial dataset, which is not equivalent to the traditional formulation of DP and has certain limitations \cite{charest2017meaning} (e.g. it only concerns a given dataset). Nevertheless, it may be useful in the situations where strict privacy guarantees are not required or cannot be achieved by existing methods, or when one wants to get a better idea about expected privacy loss rather than the highly unlikely worst-case.

Lastly, all existing limitations of GANs (or generative models in general), such as training instability or mode collapse, will apply to this method. Hence, at the current state of the field, our approach may be difficult to adapt to inputs other than image data. Yet, there is still a number of privacy-sensitive applications, e.g. medical imaging or facial analysis, that could benefit from our technique. And as generative methods progress, new uses will be possible.

\section{Evaluation}
\label{sec:evaluation}

\begin{table}
	\caption{Accuracy of student models for non-private baseline, PATE~\cite{papernot2016semi}, and our method.}
	\label{tab:accuracy}
	\centering
	\begin{tabular}{ c c c c }
		\toprule
		{\bf Dataset} 	& {\bf Non-private} & {\bf PATE} & {\bf Our approach} \\
		\midrule
		MNIST 		& $99.2\%$ 	& $98.0\%$ 	& $98.3\%$ \\
		SVHN            	& $92.8\%$ 	& $82.7\%$ 	& $87.7\%$ \\
		\bottomrule
	\end{tabular}
\end{table}

In this section, we describe the experimental setup and implementation, and evaluate our method on MNIST~\cite{lecun1998gradient}, SVHN~\cite{netzer2011reading}, and CelebA~\cite{liu2015faceattributes} datasets.

\subsection{Experimental Setting}

We evaluate our method in two major ways. First, we show that not only is it feasible to train ML models purely on generated data, but it is also possible to achieve high learning performance (Section~\ref{sec:learning}). Second, we compute empirical bounds on expected privacy loss and evaluate the effectiveness of artificial data against model inversion attacks (Section~\ref{sec:privacy}).

Learning performance experiments are set up as follows:
\begin{enumerate}
\item Train a generative model (\emph{teacher}) on the original dataset using only the training split.
\item Generate an artificial dataset by the obtained model and use it to train ML models (\emph{students}).
\item Evaluate students on a held-out test set. 
\end{enumerate}


Note that there is no dependency between teacher and student models. Moreover, student models are not constrained to neural networks and can be implemented as any type of machine learning algorithm.

We choose three commonly used image datasets for our experiments: MNIST, SVHN, and CelebA. MNIST is a handwritten digit recognition dataset consisting of 60000 training examples and 10000 test examples, each example is a 28x28 size greyscale image. SVHN is also a digit recognition task, with 73257 images for training and 26032 for testing. The examples are coloured 32x32 pixel images of house numbers from Google Street View. CelebA is a facial attributes dataset with 202599 images, each of which we crop to 128x128 and then downscale to 48x48.

\begin{table}
	\caption{Empirical privacy parameters: expected privacy loss bound $\mu$ and probability $\gamma$ of exceeding it.}
	\label{tab:privacy}
	\centering
	\begin{tabular}{l l c c c}
		\toprule
		{\bf Dataset} 			& {\bf Method}			& $\mu$ & $\gamma$ \\
		\midrule
		\multirow{2}{*}{MNIST} 	& WGAN-GP 			& $5.80$ 		&  \\
							& WGAN-GP (DP critic) 	& $5.36$ 		&  \\
		\addlinespace
		\multirow{2}{*}{SVHN} 	& WGAN-GP 			& $13.16$ 	& \multirow{2}{*}{$10^{-5}$} \\
							& WGAN-GP (DP critic) 	& $4.92$ 		& \\
		\addlinespace
		\multirow{2}{*}{CelebA} 	& WGAN-GP 			& $6.27$ 		& \\
							& WGAN-GP (DP critic) 	& $4.15$ 		& \\
		\bottomrule
	\end{tabular}
\end{table}

\subsection{Implementation Details}
\label{sec:implementation}

For our experiments, we use Python and Pytorch framework.\footnote{\url{http://pytorch.org}} We implement, with some minor modifications, a Wasserstein GAN with gradient penalty (WGAN-GP) by \citeauthor{gulrajani2017improved} \shortcite{gulrajani2017improved}. More specifically, the critic consists of four convolutional layers with SELU~\cite{klambauer2017self} activations (instead of ReLU) followed by a fully connected linear layer which outputs a $d$-dimensional feature vector ($d=64$). For the DP critic, we implement the Gaussian noise mechanism~\cite{dwork2014algorithmic} by clipping the $L2$-norm of this feature vector to $C=1$ and adding Gaussian noise with $\sigma=1.5$ (we refer to it as \emph{DP layer}). Finally, it is passed through a linear classification layer. The generator starts with a fully connected linear layer that transforms noise and labels into a $4096$-dimensional feature vector which is then passed through a SELU activation and three deconvolution layers with SELU activations. The output of the third deconvolution layer is downsampled by max pooling and normalised with a \texttt{tanh} activation function.

Similarly to the original paper, we use a classical WGAN value function with the gradient penalty that enforces Lipschitz constraint on a critic. We also set the penalty parameter $\lambda=10$ and the number of critic iterations $n_\text{critic} = 5$. Furthermore, we modify the architecture to allow for conditioning WGAN on class labels. Binarised labels are appended to the input of the generator and to the linear layer of the critic after convolutions. Therefore, the generator can be used to create labelled datasets for supervised learning.

Both networks are trained using Adam~\cite{kingma2015adam} with learning rate $10^{-4}$, $\beta_1 = 0$, $\beta_2 = 0.9$, and a batch size of $64$.

\begin{figure}
	\centering
	\includegraphics[width=\linewidth]{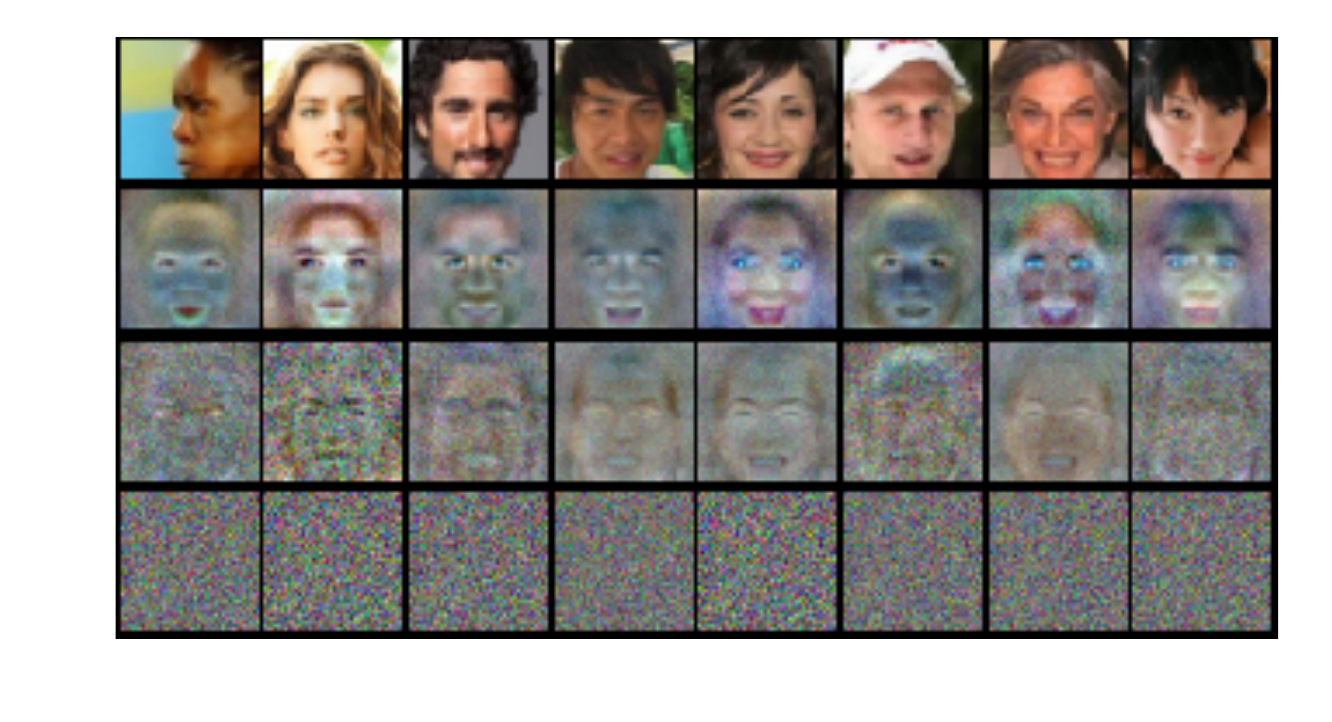}
    	\caption{Results of the model inversion attack. Top to bottom: real target images, reconstructions from non-private model, our method, and DP model.}
	\label{fig:reconstruction}
\end{figure}
\begin{table}
	\caption{Face detection and recognition rates (pairs with distances below $0.99$) for non-private, our method, and DP.}
	\label{tab:face_recognition}
	\centering
	\begin{tabular}{ c c c c }
		\toprule
		{\bf } 	& {\bf Non-private} & {\bf Our approach} & {\bf DP} \\
		\midrule
		Detection 		& $63.6\%$ 	& $1.3\%$ 	& $0.0\%$ \\
		Recognition 		& $11.0\%$ 	& $0.3\%$ 	& $-~$ \\
		\bottomrule
	\end{tabular}
\end{table}

The student network is constructed of two convolutional layers with ReLU activations, batch normalisation and max pooling, followed by two fully connected layers with ReLU, and a \texttt{softmax} output layer. Note that this network does not achieve state-of-the-art performance on the used datasets, but we are primarily interested in evaluating the relative performance drop compared to a non-private model.

To estimate privacy loss, we carry out the procedure presented in Section~\ref{sec:approach}. Specifically, based on recent ideas in image qualitative evaluation, e.g. FID and Inception Score, we compute image features by the Inception V3 network~\cite{szegedy2016rethinking} and use inverse distances between features as $sim$ function. We implement the KL divergence estimator~\cite{perez2008kullback} and use $k$-d trees~\cite{bentley1975multidimensional} for fast nearest neighbour searches. For privacy evaluation, we implement the model inversion attack.

\subsection{Learning Performance}
\label{sec:learning}

First, we evaluate the generalisation ability of a student model trained on artificial data. More specifically, we train a student model on generated data and report test classification accuracy on a held-out real set.

As noted above, most of the work on privacy-preserving ML focuses on \emph{model release} methods and assumes (explicitly or implicitly) access to similar "public" data in one form or another~\cite{abadi2016deep,papernot2016semi,papernot2018scalable,zhang2018differentially}. On the other hand, existing \emph{data release} solutions struggle with high-dimensional data~\cite{zhu2017differentially}. It limits the choice of methods for comparison.

We chose to compare learning performance with the current state-of-the-art model release technique, PATE by \citeauthor{papernot2018scalable}~\shortcite{papernot2018scalable}, which uses a relatively small set of unlabelled "public" data. Since our approach does not require any "public" data, in order to make the evaluation more appropriate, we pick the results of PATE corresponding to the least number of labelling queries.

\begin{figure}
	\centering
	\includegraphics[width=\linewidth]{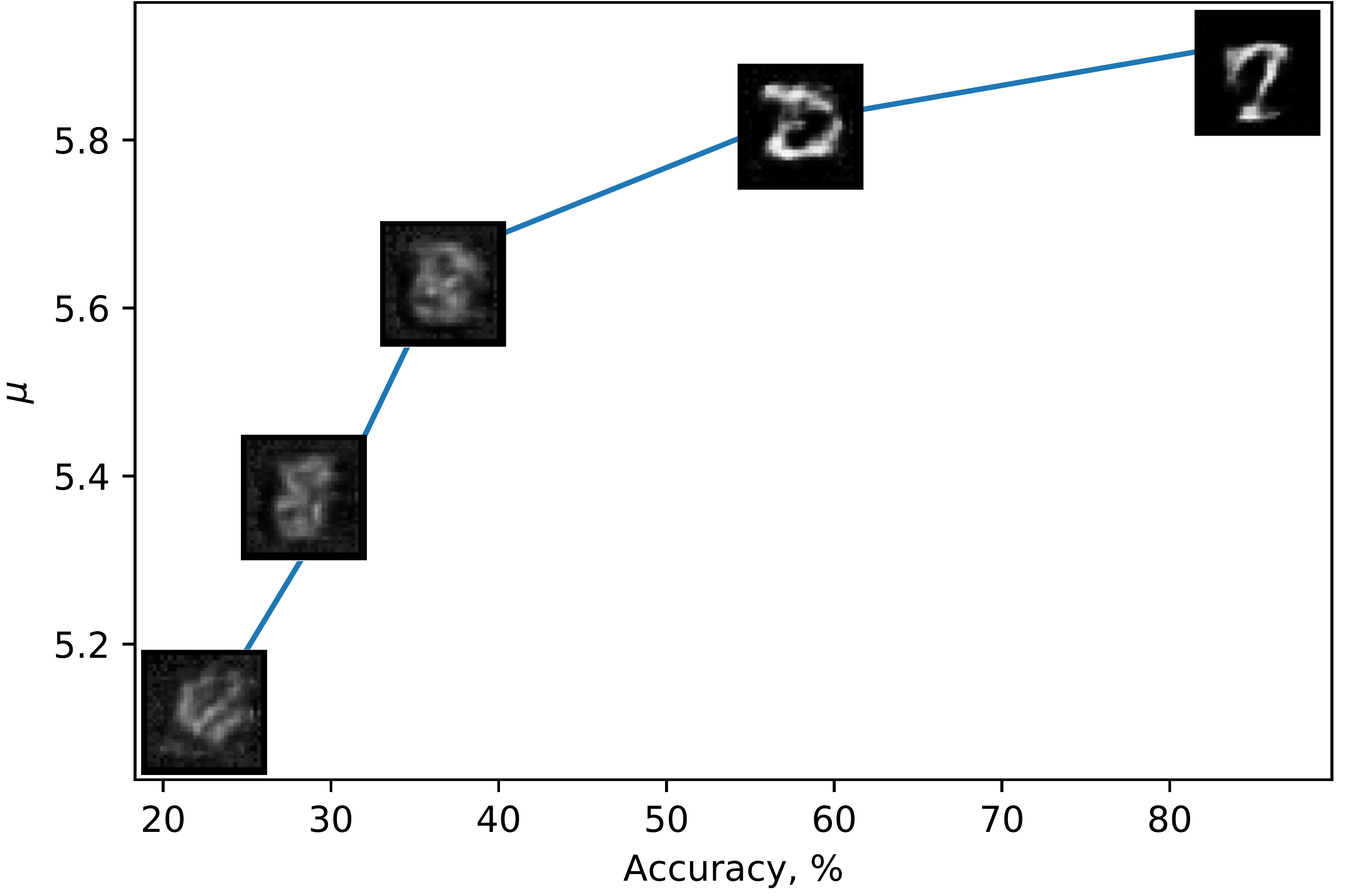}
    	\caption{Privacy-accuracy trade-off curve and corresponding image reconstructions from a multi-layer perceptron trained on artificial MNIST dataset.}
	\label{fig:privacy_accuracy_reconstruction}
\end{figure}

Table~\ref{tab:accuracy} shows test accuracy for the non-private baseline model (trained on the real training set), PATE, and our method. We observe that artificial data allows us to achieve $98.3\%$ accuracy on MNIST and $87.7\%$ accuracy on SVHN, which is comparable or better than corresponding results of PATE. These results demonstrate that our approach does not compromise learning performance, and may even improve it, while enabling the full flexibility of data release methods.

Additionally, we train a simple logistic regression model on artificial MNIST samples, and obtain $91.69\%$ accuracy, compared to $92.58\%$ on the original data, confirming that student models are not restricted to a specific type.

Furthermore, we observe that one could use artificial data for validation and hyper-parameter tuning. In our experiments, correlation coefficients between real and artificial validation losses range from 0.7197 to 0.9972 for MNIST and from 0.8047 to 0.9810 for SVHN.

\subsection{Privacy Analysis}
\label{sec:privacy}

\begin{figure}
	\centering
	\begin{subfigure}{0.49\linewidth}
    		\includegraphics[width=\textwidth]{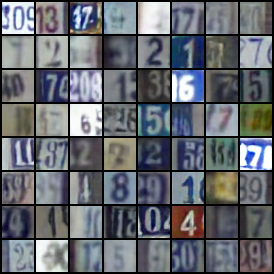}
    		\caption{Generated}
	\end{subfigure}
	\begin{subfigure}{0.49\linewidth}
    		\includegraphics[width=\textwidth]{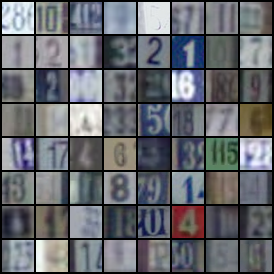}
    		\caption{Real}
	\end{subfigure}
	\caption{Generated and closest real examples for SVHN.}
	\label{fig:most_similar_svhn}
\end{figure}

Using the privacy estimation framework (see Section~\ref{sec:approach}), we fix the probability $\gamma$ of exceeding the expected privacy loss bound $\mu$ in all experiments to $10^{-5}$ and compute the corresponding $\mu$ for each dataset and two versions of WGAN-GP (vanilla and with DP critic). Table~\ref{tab:privacy} summarises our findings. It is worth noting, that our $\mu$ should not be viewed as an empirical estimation of $\varepsilon$ of DP, since the former bounds \emph{expected} privacy loss while the latter--\emph{maximum}. These two quantities, however, in our experiments turn out to be similar to deep learning DP bounds found in recent literature~\cite{abadi2016deep,papernot2018scalable}. This may be explained by tight concentration of privacy loss random variable~\cite{dwork2016concentrated} or loose estimation. Additionally, DP critic helps to bring down $\mu$ values in all cases.

The lack of theoretical privacy guarantees for our method neccesitates assessing the strength of provided protection. We perform this evaluation by running the \emph{model inversion attack}~\cite{fredrikson2015model} on a student model. Note that we also experimented with another well-known attack on machine learning models, the membership inference~\cite{shokri2017membership}. However, we did not include it in the final evaluation, because of the poor attacker's performance in our setting (nearly random guess accuracy for given datasets and models even without any protection).

In order to run the attack, we train a student model (a simple multi-layer perceptron with two hidden layers of 1000 and 300 neurons) in three settings: real data, artificial data generated by GAN (with DP critic), and real data with differential privacy (using DP-SGD with a small $\varepsilon < 1$). As facial recognition is a more privacy-sensitive application, and provides a better visualisation of the attack, we picked CelebA attribute prediction task to run this experiment.

Figure~\ref{fig:reconstruction} shows the results of the model inversion attack. The top row presents the real target images. The following rows depict reconstructed images from a non-private model, a model trained on GAN samples, and DP model, correspondingly. One can observe a clear information loss in reconstructed images going from non-private model, to artificial data, to DP. The latter is superior in decoupling the model and the training data, and is a preferred choice in the model release setting and/or if public data is accessible for pre-training. The non-private model, albeit trained with abundant data (${\sim}200K$ images) reveals facial features, such as skin and hair colour, expression, etc. Our method, despite failing to conceal general shapes in training images (i.e. faces), seems to achieve a trade-off, hiding most of the specific features. The obtained reconstructions are either very noisy (columns 1, 2, 6, 8), much like DP, or converge to some average feature-less faces (columns 4, 5, 7).

We also analyse real and reconstructed image pairs using OpenFace~\cite{amos2016openface} (see Table~\ref{tab:face_recognition}). It confirms our initial findings: in images reconstructed from a non-private model, faces were detected (recognised) $63.6\%~(11\%)$ of the time, while for our method, detection succeeded only in $1.3\%$ of cases and recognition rate was $0.3\%$, well within state-of-the-art error margins. For DP both rates were at $0\%$.

To evaluate our privacy estimation method, we look at how the privacy loss bound $\mu$ correlates with the success of the attack. Figure~\ref{fig:privacy_accuracy_reconstruction} depicts the privacy-accuracy trade-off curve for an MLP (64-32-10) trained on artificial data. In this setting, we use a stacked denoising autoencoder to compress images to 64-dimensional feature vectors and facilitate the attack performance. Along the curve, we plot examples of the model inversion reconstruction at corresponding points. We see that with growing $\mu$, meaning lower privacy, both model accuracy and reconstruction quality increase.

\begin{figure}
	\centering
	\begin{subfigure}{0.49\linewidth}
    		\includegraphics[width=\textwidth]{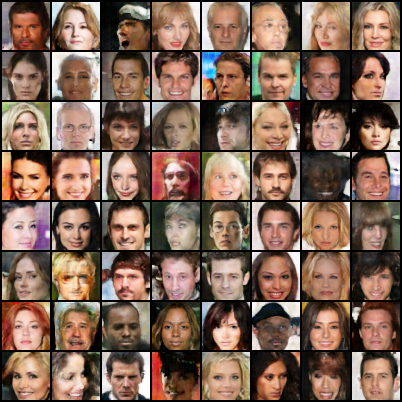}
    		\caption{Generated}
	\end{subfigure}
	\begin{subfigure}{0.49\linewidth}
    		\includegraphics[width=\textwidth]{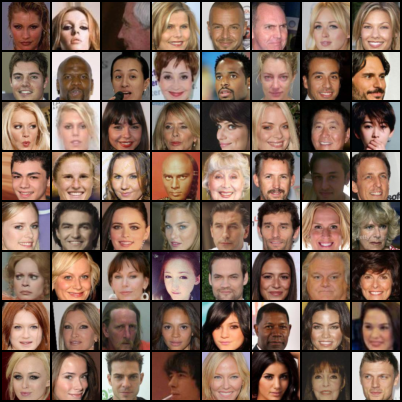}
    		\caption{Real}
	\end{subfigure}
	\caption{Generated and closest real examples for CelebA.}
	\label{fig:most_similar_celeba}
\end{figure}

Finally, as an additional measure, we perform visual inspection of generated examples and corresponding nearest neighbours in real data. Figures~\ref{fig:most_similar_svhn} and~\ref{fig:most_similar_celeba} depict generated and the corresponding most similar real images from SVHN and CelebA datasets. We observe that, despite general visual similarity, generated images differ from real examples in details, which is normally more important for privacy. For SVHN, digits vary either in shape, colour or surroundings. A lot of pairs come from different classes. For CelebA, the pose and lighting may be similar, but such details as gender, skin colour, facial features are usually significantly different.

\section{Conclusions}
\label{sec:conclusion}

We investigate the problem of private data release for complex high-dimensional data. In contrast to commonly studied model release setting, this approach enables important advantages and applications, such as data pooling from multiple sources, simpler development process, and data trading. 

We employ generative adversarial networks to produce artificial privacy-preserving datasets. The choice of GANs as a generative model ensures scalability and makes the technique suitable for real-world data with complex structure. Unlike many prior approaches, our method does not assume access to similar publicly available data.
In our experiments, we show that student models trained on artificial data can achieve high accuracy on MNIST and SVHN datasets. Moreover, models can also be validated on artificial data.

We propose a novel technique for estimating privacy of released data by empirical bounds on expected privacy loss. We compute privacy bounds for samples from WGAN-GP on MNIST, SVHN, and CelebA, and demonstrate that expected privacy loss is bounded by single-digit values. To evaluate provided protection, we run a model inversion attack and show that training with GAN reduces information leakage (e.g. face detection drops from $63.6\%$ to $1.3\%$) and that attack success correlates with estimated privacy bounds.

Additionally, we introduce a simple modification to the critic: differential privacy layer. Not only does it improve privacy loss bounds and ensures DP guarantees for the critic output, but it also acts as a regulariser, improving stability of training, and quality and diversity of generated images.

Considering the rising importance of privacy research and the lack of good solutions for private data publishing, there is a lot of potential future work. In particular, a major direction of advancing current work would be achieving differential privacy guarantees for generative models while still preserving high utility of generated data. A step in another direction would be to improve the privacy estimation framework, e.g. by bounding maximum privacy loss, or finding a more principled way of sampling from outcome distributions.

\bibliographystyle{aaai}
\bibliography{pal_2019}

\end{document}